\documentclass[a4paper,fleqn]{cas-sc}

\usepackage[numbers]{natbib}
\usepackage{makecell}
\usepackage{booktabs}
\usepackage{color, colortbl}
\usepackage{array}
\usepackage{longtable}
\usepackage{multirow}
\usepackage{nccmath}
\usepackage{appendix}
\usepackage{pifont}

\definecolor{Gray}{gray}{0.9}
\newcommand{\ie}{\emph{i.e.},~}
\newcommand{\eg}{\emph{e.g.},~}
\newcommand{\ea}{\emph{et al.}~}
\newcommand{\vs}{\emph{vs.}~}
\newcommand{\bsage}{BSAGE~}
\def\tsc#1{\csdef{#1}{\textsc{\lowercase{#1}}\xspace}}
\tsc{WGM}
\tsc{QE}

\begin{document}
\let\WriteBookmarks\relax
\def\floatpagepagefraction{1}
\def\textpagefraction{.001}

\shorttitle{Brain Structure Ages - A new biomarker for multi-disease classification}    

\shortauthors{HD Nguyen}  

\title [mode = title]{Brain Structure Ages - A new biomarker for multi-disease classification}

%

\author[1]{Huy-Dung Nguyen}[orcid=0000-0002-3980-8029]
\cormark[1]

\ead{huy-dung.nguyen@u-bordeaux.fr}

\author[1]{Michaël Clément}

\author[1]{Boris Mansencal}

\author[1]{Pierrick Coupé}

\affiliation[1]{organization={Univ. Bordeaux, CNRS, Bordeaux INP, LaBRI, UMR 5800},
    postcode={33400 Talence},
    country={France}}

\cortext[1]{Corresponding author}

\begin{abstract}
Age is an important variable to describe the expected brain's anatomy status across the normal aging trajectory. The deviation from that normative aging trajectory may provide some insights into neurological diseases. In neuroimaging, predicted brain age is widely used to analyze different diseases. However, using only the brain age gap information (\ie the difference between the  chronological age and the estimated age) can be not enough informative for disease classification problems. In this paper, we propose to extend the notion of global brain age by estimating brain structure ages using structural magnetic resonance imaging. To this end, an ensemble of deep learning models is first used to estimate a 3D aging map (\ie voxel-wise age estimation). Then, a 3D segmentation mask is used to obtain the final brain structure ages. This biomarker can be used in several situations. First, it enables to accurately estimate the brain age for the purpose of anomaly detection at the population level. In this situation, our approach outperforms several state-of-the-art methods. Second, brain structure ages can be used to compute the deviation from the normal aging process of each brain structure. This feature can be used in a multi-disease classification task for an accurate differential diagnosis at the subject level. Finally, the brain structure age deviations of individuals can be visualized, providing some insights about brain abnormality and helping clinicians in real medical contexts.
\end{abstract}




\maketitle

\section{Introduction}
In the medical field, chronological age is widely used as an indicator to describe people. It depicts a reference curve that healthy organs should follow. The deviation from that reference may be associated with different factors such as the interaction of genes, environment, lifestyle and diseases \cite{franke_ten_2019}. To measure this deviation, the concept of biological age (BA) has been created. It is an estimation of individual's age based on various advanced strategies \cite{chang_electrocardiogram-based_2022, raghu_deep_2021, nakamura_method_2007} and is expected to be able to take into account all the factors mentioned above. Consequently, an accelerated (or delayed) aging process results in a higher (or lower) value of BA with respect to the chronological age.

The analysis of BA can be associated with a whole-body system or a specific organ. On the one hand, the whole-body evaluation approaches typically use non-imaging data (\eg DNA methylation patterns \cite{chen_dna_2016}, protein \cite{ignjatovic_age-related_2011}) and are unable to account for the variations in aging between individual organs \cite{armanious_age-net_2021}. Such global information might be difficult to use in clinical practice. On the other hand, imaging studies of BA dedicated to a particular organ may provide important details about that organ's condition, and the brain is one of the most commonly studied organs. Brain structure changes are demonstrated to be mutually caused by the natural aging process and neurodegenerative diseases \cite{peters_aging_2006, huizinga_spatio-temporal_2018, jia_longitudinal_2015, tisserand_voxel-based_2004, coupe_lifespan_2019, coupe_hippocampalamygdaloventricular_2022}. Cole \ea demonstrated that biological brain age can enable the development of treatment plans and a better understanding of disease processes \cite{cole_predicting_2017}. The authors emphasized that the difference between the predicted brain age and the chronological age is a valuable bio-marker since it shows a correlation with aging as well as with diseases. This difference is denoted as BrainAGE for Brain Age Gap Estimation. Since its introduction, this new bio-marker has been widely used in many studies to analyze various diseases \cite{franke_ten_2019}. Generally, a model is trained with brain images from a healthy population and then used to estimate the age of patients with diseases.

In BrainAGE, structural magnetic resonance imaging (sMRI) is the most used modality (about 88\% of studies \cite{mishra_review_2021}). It has been shown that reasonable prediction error can be achieved using this modality. Moreover, sMRI is commonly available in medical environments \cite{mishra_review_2021}. Initially, sMRI was used with some traditional machine learning algorithms such as relevance vector regression \cite{franke_estimating_2010}, support vector regression \cite{liem_predicting_2017} and Gaussian process regression \cite{cole_brain_2018} to perform BrainAGE. The prediction error of these methods ranges from 4.29 to 5.02 years for the mean absolute error (MAE) metric. Since the success of deep learning in many natural image processing applications, it has also become a useful technique in various medical imaging studies. Recent studies show the capacity of deep learning algorithms in the brain age estimation task based on sMRI with an MAE ranging from 1.96 to 4.16 years \cite{armanious_age-net_2021, bintsi2020patch, cole_predicting_brain_2017, jonsson_brain_2019, bermudez_anatomical_2019}. These promising results suggest using deep learning to estimate brain age for further analysis.

These deep learning based methods adapt famous convolutional neural network (CNN) architectures to estimate the brain age. When employing a VGG-like architecture, Ueda \ea demonstrated that using 3D CNN can lead to better accuracy than 2D CNN for age prediction \cite{ueda_age_2019}. In another work, Cole \ea also used a VGG-like architecture and found that the grey matter extracted from 3D sMRI is better than white matter and raw image for age prediction \cite{cole_predicting_2017}. Using a similar architecture, Bermudez \ea suggested to additionally take advantage of brain structure volume to improve the model performance \cite{bermudez_anatomical_2019}. Bintsi \ea employed ResNet architecture to predict age on several sub-volumes of brain image \cite{kia_patch-based_2020}. The final prediction was aggregated using a linear regression model. Armanious \ea proposed to use the inception module with squeeze-and-excitation module to accurately predict healthy brain age \cite{armanious_age-net_2021}. Bashyam \ea customized the inception-resnetv2 to build their model and trained it on 11729 healthy subjects.

After training a brain age prediction model, the next step is to apply it to a population of interest to compare healthy and diseased groups (\ie analysis at population level). For example, Franke \ea analyzed the brain maturation during childhood and adolescence \cite{franke_brain_2012}. By applying a trained model on subjects being born before the 28th and after the 29th week of gestation, they found that the BrainAGE of the first group was significantly lower than the second group, showing a delayed structural brain maturation of the first group. Applying the same technique, Koutsouleris \ea demonstrated an accelerated aging of 5.5 years in schizophrenia and 4.0 years in major depression patients compared to normal aging \cite{koutsouleris_accelerated_2014}. In another study dedicated to Alzheimer's Disease (AD), the BrainAGE was estimated about +10 years in AD patients, implying accelerated aging of this population \cite{franke_estimating_2010}.

Although the BrainAGE can provide a description of a specific population, its application in individual diagnosis is still limited. Only a few works suggested performing disease detection or differential diagnosis using BrainAGE at subject level. For instance, the BrainAGE was used as a biomarker to perform differential diagnosis between mild cognitive impairment and AD in \cite{gaser_brainage_2013} and to diagnose AD (\ie AD patients \vs healthy controls) in \cite{franke2014dementia, varzandian_classification-biased_2021}. More recently, Cheng \ea used deep learning to accurately predict brain age and they use BrainAGE as the only feature for various binary diagnosis tasks (\ie diseased subjects \vs healthy subjects) \cite{cheng_brain_2021}. Although encouraging results were obtained, these works performed only binary classification tasks but not multi-class classification. The reason for this may be due to the coarse description of brain's state provided by the global BrainAGE. Indeed, BrainAGE can only describe the aging process of the whole brain but does not provide any details about brain structures' state. Therefore, it is difficult to use BrainAGE for involved tasks such as the differential diagnosis of multiple pathologies.

In this paper, we propose to extend the notion of the global brain age to local brain structure ages. Our main hypothesis is that the aging process is heterogeneous over the brain and specifically, different brain structures may present different ages. Consequently, we first estimate the brain age at the voxel level. This results in a 3D aging map of voxelwise brain ages. By averaging predicted brain ages by brain structure, we obtain the Brain Structure Ages, denoted as BSA. This local BSA is expected to provide more information about the subject's condition than a global age prediction of a whole subject's brain. As shown later, this novel biomarker can be used as input of a multi-layer perceptron (MLP) to accurately estimate the subject's age. During validation, our framework showed competitive results compared to state-of-the-art methods. Furthermore, the difference between BSA and the subject's chronological age, denoted as \bsage for Brain Structure Age Gap Estimation, can be also used with a support vector machine classifier (SVM) for multi-class classification (\ie Cognitively Normal (CN) \vs AD \vs Frontotemporal Disease (FTD) \vs Multiple Sclerosis (MS) \vs Parkinson's disease (PD) \vs Schizophrenia (SZ)). In our experiments, we demonstrated the important gain of using \bsage compared to BrainAGE for the multi-disease classification task. Finally, by projecting the \bsage on a brain atlas, we can visually observe the brain regions affected by different diseases.
\section{Materials}

\subsection{Datasets}
The data used in this study comprise 39255 images from various datasets: the Autism Brain Imaging Data Exchange (ABIDE) \cite{di_martino_autism_2014, di_martino_enhancing_2017}, the Alzheimer’s Disease Neuroimaging Initiative (ADNI) \cite{jack_alzheimers_2008}, the Australian Imaging Biomarkers and Lifestyle Study of aging (AIBL) \cite{ellis_australian_2009}, the International Consortium for Brain Mapping (ICBM) \cite{Mazziotta_icbm}, the Information eXtraction from Images (IXI) \footnote{\url{https://brain-development.org/ixi-dataset/}}, the National Database for Autism Research (NDAR) \cite{payakachat_national_2016}, the Open Access Series of Imaging Studies (OASIS) \cite{lamontagne_oasis-3_2019}, Cincinnati MR Imaging of Neurodevelopment (C-MIND) \footnote{\url{https://nda.nih.gov/edit_collection.html?id=2329}}, UKBioBank \cite{bycroft_uk_2018}, the Strategic Research Program for Brain Sciences (SRPBS) \cite{tanaka_multi-site_2021}, the Center for Biomedical Research Excellence (COBRE) \footnote{\url{http://fcon_1000.projects.nitrc.org/indi/retro/cobre.html}}, the Cambridge Centre for aging and Neuroscience (CamCAN) \cite{taylor_cambridge_2017}, the Parkinson’s Progression Markers Initiative (PPMI) \cite{marek_parkinson_2011}, the Frontotemporal Lobar Degeneration Neuroimaging Initiative (NIFD) \footnote{\url{https://ida.loni.usc.edu/collaboration/access/appLicense.jsp}}, the Observatoire Français de la Sclérose en Plaques (OFSEP) \cite{vukusic_observatoire_2020}, the National Alzheimer's Coordinating Center (NACC) \cite{beekly_national_2007}, the Dallas Lifespan Brain Study (DLBS) \footnote{\url{https://fcon_1000.projects.nitrc.org/indi/retro/dlbs.html}}, the Minimal Interval Resonance Imaging in Alzheimer's Disease \cite{malone_miriadpublic_2013}, the Minimal Interval Resonance Imaging in Alzheimer’s Disease (MIRIAD) \cite{malone_miriadpublic_2013}, and a study on schizophrenia (BrainGluSchi) \cite{bustillo_glutamatergic_2016}. All the T1 weighted images at the baseline were used.

\subsubsection{Chronological age prediction}
Among available data, 32718 images were used to study the accuracy of our chronological age predictor. First, eight datasets including 2887 images (\ie ABIDE~I, ADNI, AIBL, ICBM, C-MIND, IXI, NDAR, OASIS1) were used in training/validation. Second, two external datasets (\ie out-of-domain) were used for testing. Concretely, CN subjects of ABIDE~II (\ie 580 images) were used to estimate the model accuracy on a young population and CN from UKBioBank (\ie 29251 images) were used to estimate the model accuracy on an older population (see Table~\ref{table:data_age_predictor}). For ABIDE, we ensured that no subject in phase I was presented in phase II.

\subsubsection{Multiple pathologies classification}
Besides, we assessed the classification performance using \bsage on 6537 images composed of 6 classes (\ie CN, AD, FTD, MS, PD and SZ). Eight datasets including 1992 images (ADNI, AIBL, SRPBS, COBRE, CamCAN, PPMI~phase~1, NIFD and OFSEP~centers~1-2) were used to perform a 10-fold cross validation (in-domain validation) (see Table \ref{table:data_classification}). Then, we constructed an out-of-domain dataset including 4545 images using seven cohorts (\ie NACC, DLBS, MIRIAD, OASIS3, BrainGluShi, PPMI~phase~2 and OFSEP-other-centers) to assess the generalization capacity of such models. For the OFSEP, we used the acquisition sites to split this global dataset into two non-overlapping domains. For PPMI, we ensured that no subject in phase I was presented in phase II.

\begin{table}[htbp]
\begin{minipage}{\textwidth}
\begin{center}
\caption{On top, summary of participants used for training age predictor. On bottom, description of the external datasets used for testing. }\label{table:data_age_predictor}
\begin{tabular*}{0.8\textwidth}{@{\extracolsep{\fill}}cccc}
\toprule
Usage & Dataset & Male/Female& Age (Mean ± Std)\\

\midrule
\multirow{9}{*}{\makecell{Age prediction \\ training}} & ABIDE~I & 408/84 & 17.5 ± 7.8\\
\cmidrule{2-4}
& ADNI & 201/203 & 74.8 ± 5.8\\
\cmidrule{2-4}
& AIBL & 112/120 & 72.3 ± 6.7\\
\cmidrule{2-4}
& ICBM & 112/182 & 33.7 ± 14.3\\
\cmidrule{2-4}
& C-MIND & 107/129 & 8.4 ± 4.3\\
\cmidrule{2-4}
& IXI & 242/307 & 48.8 ± 16.5\\
\cmidrule{2-4}
& NDAR & 208/174 & 12.4 ± 6.0\\
\cmidrule{2-4}
& OASIS1 & 111/187 & 45.3 ± 23.8\\
\midrule

\makecell{Young population \\ testing} & ABIDE~II & 403/177 & 14.8 ± 9.3\\
\midrule

\makecell{Older population \\ testing} & UKBioBank & 14917/14334 & 64.2 ± 7.9\\
\midrule
& \textbf{Total} & 16821/15897 & 14.8 ± 9.3\\

\bottomrule
\end{tabular*}
\end{center}
\end{minipage}
\end{table}

\begin{table}[htbp]
\caption{Number of participants (Male/Female) used for multi-class classification.}\label{table:data_classification}
\begin{center}
\begin{minipage}{\textwidth}
\begin{tabular*}{\textwidth}{@{\extracolsep{\fill}}lccccccc}
\toprule
\textbf{Usage} & \textbf{Dataset} & \textbf{CN} & \textbf{AD} & \textbf{FTD} & \textbf{MS} & \textbf{PD} & \textbf{SZ} \\

\midrule

& ADNI &  & 181/150 \\
\cmidrule{2-8}

& AIBL &  & 18/28 \\
\cmidrule{2-8}

\multirow{8}{*}{\makecell[l]{10-fold\\cross\\validation\\training}}& SRPBS & 88/60 & & & & & 84/58\\
\cmidrule{2-8}

& COBRE & 11/7 & & & & & 54/14\\
\cmidrule{2-8}

& CamCAN & 75/85\\
\cmidrule{2-8}

& \makecell{PPMI\\phase 1} & 35/13 & & & & 228/131\\
\cmidrule{2-8}

& NIFD & 15/15 & & 87/56\\
\cmidrule{2-8}

& \makecell{OFSEP\\centers 1-2} & & & & 161/338\\
\midrule

& NACC & 47/104 & 318/419 & 22/23\\
\cmidrule{2-8}

& DLBS & 117/196\\
\cmidrule{2-8}

\multirow{7}{*}{\makecell[l]{Out-of\\-domain\\ testing}}& MIRIAD & 12/11 & 19/27\\
\cmidrule{2-8}

& OASIS3 & 270/385 & 46/46\\
\cmidrule{2-8}

& \makecell{Brain-\\GluSchi} & 61/25 & & & & & 71/11\\
\cmidrule{2-8}

& \makecell{PPMI\\phase 2} & & & & & 74/58\\
\cmidrule{2-8}

& \makecell{OFSEP\\other\\centers} & & & & 585/1598\\

\midrule

& \textbf{Total} & 731/901 & 582/670 & 109/79 & 746/1936 & 302/189 & 209/83\\
\bottomrule
\end{tabular*}
\end{minipage}
\end{center}
\end{table}

\begin{figure}[ht]
\centering
\includegraphics[width=\textwidth]{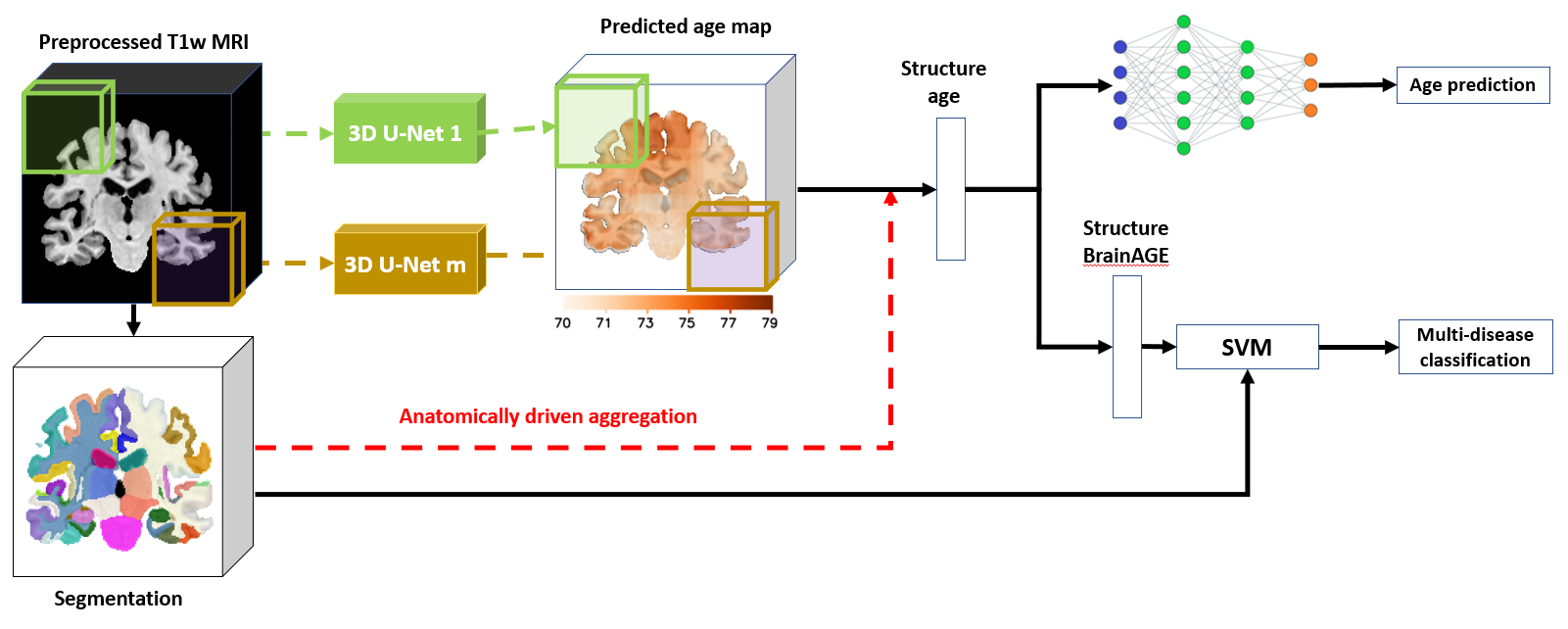}
\caption{An overview of the proposed method. The T1w image, its segmentation and the age map are taken from a 71 years old healthy person.}
\label{figure:pipeline}
\end{figure}
\subsection{Preprocessing}
The preprocessing procedure is composed of five steps: (1) denoising image \cite{manjon_adaptive_2010}, (2) inhomogeneity correction \cite{tustison_n4itk_2010}, (3) affine registration into the MNI152 space ($181\times217\times181$ voxels at $1mm\times1mm\times1mm$) \cite{avants_reproducible_2011}, (4) intensity standardization \cite{manjon_robust_2008} and (5) intracranial cavity (ICC) extraction \cite{manjon_nonlocal_2014}. After preprocessing, we used AssemblyNet~\footnote{Available at \url{https://github.com/volBrain/AssemblyNet}}  \cite{coupe_assemblynet_2020} to get the parcellation of the brain into 133 structures (see Figure \ref{figure:pipeline}). This brain structure segmentation is then used to compute the BSA for further analysis.
\section{Method}
\label{section:method}

\subsection{Method overview}
Figure \ref{figure:pipeline} provides an overview of our method. First, we estimate the brain ages map at voxel level from a preprocessed T1 image using a large number of U-Nets. Then, this 3D map is used with a segmentation mask to compute the BSA features (Section \ref{section:brain_structure_age_estimation}). Finally, the BSA features can be employed to estimate the chronological age using a MLP model or combined with brain structure volumes to perform multi-disease classification using an SVM classifier (Section~\ref{section:application_to_chronological_age_prediction_and_multi_disease_classification}).
\begin{figure}
    \centering
    \includegraphics[width=0.7\textwidth]{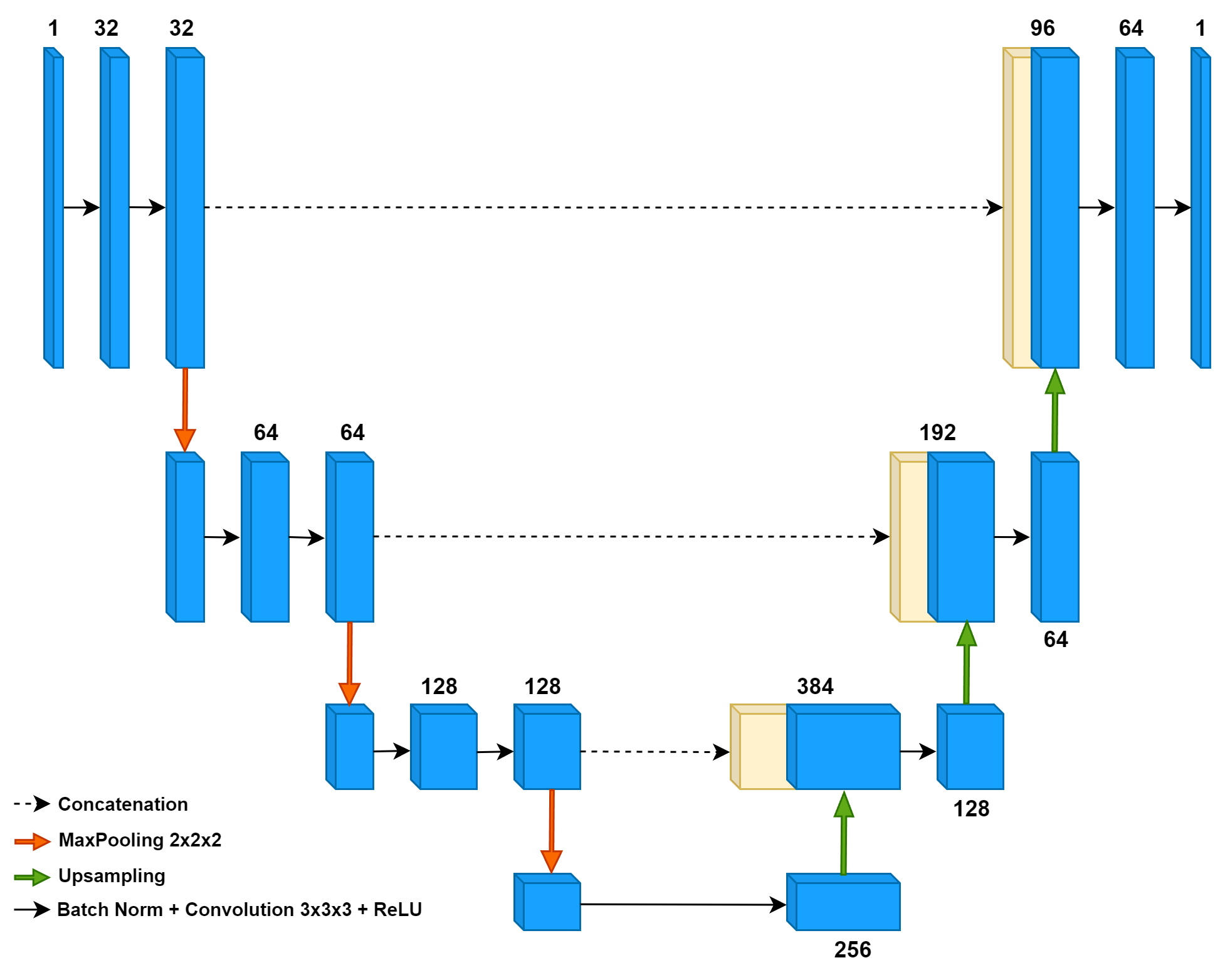}
    \caption{Architecture of an unit U-Net used for voxel-level age prediction. The number above each block is the number of channel.}
    \label{fig:UNet_architecture}
\end{figure}
\subsubsection{Brain structure age estimation}
\label{section:brain_structure_age_estimation}
In order to produce the 3D aging map, we extracted $m=k^3$ overlapping 3D sub-volumes of the same size for each T1w MRI. Next, we trained $m$ U-Nets to predict age at voxel level with these $m$ 3D sub-volumes. The goal of this training strategy is dual. First, as the size of a sub-volume is relatively small compared to the original image, it can be trained with a lighter weight model and thus, require only a low computation capacity. Second, we limit the receptive field of each model to a local brain region in order to force it to locally describe the brain age. The outputs were then used to reconstruct a 3D brain age map. Finally, the BSA was computed with the help of an AssemblyNet-based brain segmentation \cite{coupe_assemblynet_2020}. In practice, we estimated the mean value of voxel-wise age estimation for each structure segmentation.
\subsubsection{Application to chronological age prediction and multi-disease classification}
\label{section:application_to_chronological_age_prediction_and_multi_disease_classification}
To demonstrate different use cases of the BSA, we performed two experiments using this biomarker: chronological age prediction which can help to briefly describe a population and multi-disease classification which can guide clinicians to focus on certain pathologies. 

To predict the chronological age of healthy people, we employed a classical MLP and used the predicted BSA as its input. For the multi-disease classification, we first computed the \bsage (\ie the difference between BSA and the subject's chronological age) and then used it as input of an SVM classifier to address the 6-class problem CN \vs AD \vs FTD \vs MS \vs PD \vs SZ. Moreover, structure volume is used as additional feature of \bsage for SVM-based classification.
\subsection{Implementation details}
\label{section:implementation_detail}
First, a preprocessed T1w MRI in the MNI space of size $181 \times 217 \times 181$ voxels at $1mm^3$ was downscaled with a factor of 2 to the size of $91 \times 109 \times 91$ voxels. After that, we extract $k^3$ (\ie $k=5$) overlapping sub-volumes of the same size $32 \times 48 \times 32$ voxels and evenly distributed along the 3 image's dimensions from the downscale image. We trained $m=k^3$ (\ie $m=125$) U-Nets to predict age at voxel level with these $m$ sub-volumes. Figure \ref{fig:UNet_architecture} shows the architecture of our unit U-Net used for voxel-level age prediction. The $m$ outputs were then used to reconstruct a 3D age map of size $91 \times 109 \times 91$ voxels. Of note, the predicted brain age located at overlapping voxel positions of more than 1 sub-volume was averaged. The reconstructed image was upscaled using trilinear interpolation to the same spatial size as the original input. This 3D map was used to compute BSA and then \bsage features.

To train the 125 U-Nets, we use the mean absolute error (MAE) as loss function and SGD optimizer. The batch size is set to 8 and the training was terminated after 20 epochs without any improvement on validation loss. The first U-Net was trained from scratch and other U-Nets were trained with transfer learning from their adjacent U-Net (see \cite{coupe_assemblynet_2020} for more details). The training data is split into training/validation sets with a ratio of 80\%/20\% (see Table \ref{table:data_age_predictor}). In addition, when a new U-Net was trained, the training and validation data were gathered and re-split to exploit the maximum information from available data. Finally, we employed different data augmentation techniques to alleviate the overfitting problem. Concretely, we randomly shifted a patch by $t \in \{-1, 0, 1\}$ voxel in each dimension (denoted as random shift technique) and then applyed mixup data augmentation \cite{zhang_mixup_2018}.

Before using BSA features, we applied an age correction technique for each of their elements. We followed a simple method of Smith \ea \cite{smith_estimation_2019} to eliminate bias in each structure brain age. Concretely, we denoted the actual age as Y (an $N_{subjects} \times 1$ vector), the brain age as $Y_B$, a brain structure age as $X_s$ (an $N_{subjects} \times 1$ vector) and the bias term $\delta$. So, we predicted 
$Y_B$ from $X_s$: $Y_B = Y + \delta = X_s \beta$. This is equivalent to $Y =X_s \beta - \delta$. This regression can be solved with $\beta = (X_s^TX)^{-1}X_s^TY$. Finally, $Y_B = X_s(X_s^TX)^{-1}X_s^TY$.

The MLP used to estimate chronological age is composed of 4 layers with respectively $s \times 4$, $s \times 2$, $s$ and 1 neuron. To train the MLP, we used the MAE as loss function and Adam optimizer. The batch size is set to 8 and the training was terminated after 50 epochs without any improvement on validation loss.

When training the SVM for multi-disease classification, three kernels were used to select the best model through our cross-validation: linear, polynomial and radial basis function. One hundred values of C in the log-space $[-1.5;0.5]$ were used in the hyper-parameter search. We performed a grid-search for the kernel and the hyper-parameter C.

\subsection{Validation Framework}
For the chronological age prediction, we compute the BSA features of U-Nets' training subjects. This data is used to train the MLP-based regression (10-fold cross-validation). This results in 10 MLP models. At testing time, the outputs of the 10 MLP models were averaged to make the final prediction. We used two separate out-of-domain datasets to assess our method accuracy (see Table \ref{table:data_age_predictor}).

For the multi-disease classification, we also performed a 10-fold cross-validation to train the SVM classifier (see Table~\ref{table:data_classification}). We denote the classification performance on this dataset as in-domain performance. In addition, we assessed the generalization capacity on an out-of-domain dataset (see Table~\ref{table:data_classification}). Similar to the chronological age prediction problem, the outputs of 10 models were averaged to get the final prediction on these external dataset.
\section{Experimental results}
\subsection{Chronological age estimation}
\subsubsection{Ablation study}
In this part, we aim at studying different factors influencing the model performance: Data amount, different augmentation strategies (\eg random shift, mixup~\cite{zhang_mixup_2018}) and an age correction technique (see Section \ref{section:implementation_detail}). Table \ref{table:ablation_choronolical_age_pred} shows the comparison results.

First, we can observe that increasing the data amount (exp. 1, 2, 3) consistently improve the model accuracy on both young and old population in all metrics (\ie MAE and $R^2$). Second, applying different data augmentation techniques (\ie random shift, mixup in exp. 4, 5) is important in the application of age prediction. This is in line with the finding of \cite{peng_accurate_2021}. Finally, many studies have shown the advantages of using age correction techniques. In our case, the implemented technique only slightly improves the result (exp. 6). However, this technique can enhance the discriminative capacity of BSA for better disease classification (see Section \ref{section:ablation_binary_classification}). Overall, each factor contributes to our model accuracy. In the rest of the paper, BSA is computed using 100\% data, random shift, mixup and structural age correction techniques unless otherwise specified.

\begin{table}[htbp]
\caption{Ablation study for the chronological age estimation. {\color{red} Red}: best result, {\color{blue} Blue}: second best result. Text or symbols in black: Changes compared to the previous experiment. Text or symbols in gray: No change compared to the previous experiment. The model performance is estimated by different metrics: Mean absolute error (MAE) and the coefficient of determination~($R^2$).}\label{table:ablation_choronolical_age_pred}
\begin{center}
\begin{minipage}{0.8\textwidth}
\begin{tabular*}{\textwidth}{@{\extracolsep{\fill}}ccccccccc}
\toprule
No. & \rotatebox[origin=l]{90}{Data amount} & \rotatebox[origin=l]{90}{Random Shift} & \rotatebox[origin=l]{90}{MixUp} & \rotatebox[origin=l]{90}{\parbox{2cm}{Structural Age\\Correction}} & \multicolumn{2}{c}{Young population} & \multicolumn{2}{c}{Old population} \\
& & & & & MAE \raisebox{-.3\height}{\includegraphics[height=8pt]{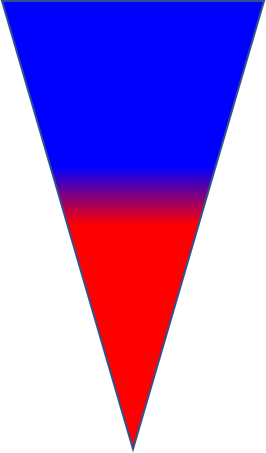}} & $R^2$ \raisebox{-.3\height}{\includegraphics[height=8pt]{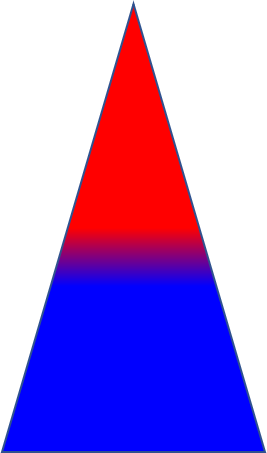}} & MAE \raisebox{-.3\height}{\includegraphics[height=8pt]{figures/arrow_down.png}} & $R^2$ \raisebox{-.3\height}{\includegraphics[height=8pt]{figures/arrow_up.png}}\\

\midrule

1 & {\color{black!40}50\%} & {\color{black!40}\ding{55}} & {\color{black!40}\ding{55}} & {\color{black!40}\ding{55}} & 4.65 & 0.30 & 8.32 & -0.89\\
2 & 75\% & {\color{black!40}\ding{55}} & {\color{black!40}\ding{55}} & {\color{black!40}\ding{55}} & 3.44 & 0.64 & 7.72 & -0.62\\
3 & 100\% & {\color{black!40}\ding{55}} & {\color{black!40}\ding{55}} & {\color{black!40}\ding{55}} & 2.38 & 0.85 & 4.60 & 0.45\\
4 & {\color{black!40}100\%} & \ding{51} & {\color{black!40}\ding{55}} & {\color{black!40}\ding{55}} & 2.11 & {\color{blue}0.89} & 3.98 & 0.58\\
5 & {\color{black!40}100\%} & {\color{black!40}\ding{51}} & \ding{51} & {\color{black!40}\ding{55}} & {\color{blue}1.92} & {\color{red}0.91} & {\color{blue}3.89} & {\color{blue}0.60}\\

\midrule
6 & {\color{black!40}100\%} & {\color{black!40}\ding{51}} & {\color{black!40}\ding{51}} & \ding{51} & {\color{red}1.91} & {\color{red}0.91} & {\color{red}3.87} & {\color{red}0.61}\\

\bottomrule
\end{tabular*}
\end{minipage}
\end{center}
\end{table}

\subsubsection{Comparison with state-of-the-art methods}
In this part, we compare our method with different state-of-the-art methods. For each method below, we used the code available \footnote{\url{https://github.com/ha-ha-ha-han/UKBiobank_deep_pretrain}} \footnote{\url{https://github.com/benniatli/BrainAgePredictionResNet}} and retrained the model using the same data split as in our training process. The first method by Jonsson \ea uses a ResNet-like architecture and demonstrated promising results in age prediction \cite{jonsson_brain_2019}. More recently, Peng \ea presented a lightweight architecture named Simple Fully Convolutional Network (SFCN) for this problem \cite{peng_accurate_2021}. They considered age prediction as a classification problem. To introduce a relationship between close classes, they used a soft label during training. The soft label is a probability distribution centered around the ground-truth age. In another work, Leonardsen \ea reused the SFCN backbone and demonstrated that the soft label can lead to better accuracy with in-domain data but the regression version presents a better generalization capacity on out-of-domain data \cite{leonardsen_deep_2022}. Table \ref{table:chronolical_pred_sota} shows the results of the comparison. For the young population, we can remark that our method presents a very low MAE (1.91 years) and very high $R^2$ (0.91) compared to other state-of-the-art methods. For the older population, all methods present a drop in performance. In this case, our method shows $MAE=3.87$ years and $R^2 = 0.61$, presenting the best prediction error over all methods.

\begin{table}[htpb]
\begin{minipage}{\textwidth}
\caption{Comparison with state-of-the-art methods. {\color{red} Red}: best result, {\color{blue} Blue}: second best result. The model performance is estimated by different metric: Mean absolute error (MAE) and the coefficient of determination ($R^2$). The results are the average accuracy of 10-fold cross validation. The age for each population is under the form: mean ± std.}\label{table:chronolical_pred_sota}
\begin{center}
\begin{tabular*}{0.8\textwidth}{@{\extracolsep{\fill}}cccccc}
\toprule
\multirow{3}{*}{No.} & \multirow{3}{*}{Method} & \multicolumn{2}{c}{Young population} & \multicolumn{2}{c}{Older population} \\
& & \multicolumn{2}{c}{Age: 14.8 ± 9.3} & \multicolumn{2}{c}{Age: 64.2 ± 7.9}\\
\cmidrule{3-6}
& & MAE \raisebox{-.3\height}{\includegraphics[height=8pt]{figures/arrow_down.png}} & $R^2$ \raisebox{-.3\height}{\includegraphics[height=8pt]{figures/arrow_up.png}} & MAE \raisebox{-.3\height}{\includegraphics[height=8pt]{figures/arrow_down.png}} & $R^2$ \raisebox{-.3\height}{\includegraphics[height=8pt]{figures/arrow_up.png}} \\
\midrule

1 & ResNet-like \cite{jonsson_brain_2019} & 2.86 & {\color{blue}0.71} & {\color{blue}4.14} & {\color{blue}0.54} \\
2 & SFCN soft label \cite{peng_accurate_2021} & {\color{blue}2.78} & {\color{blue}0.71} & 5.12 & 0.32 \\
3 & SFCN regression \cite{leonardsen_deep_2022} & 2.87 & 0.69 & 4.88 & 0.40 \\
\midrule
4 & Our method & {\color{red} 1.91} & {\color{red} 0.91} & {\color{red} 3.87} & {\color{red} 0.61}\\

\bottomrule
\end{tabular*}
\end{center}
\end{minipage}
\end{table}
\subsection{Disease classification}
\subsubsection{Ablation study for binary classification tasks}
\label{section:ablation_binary_classification}

In this part, we aim at assessing the \bsage (\ie the difference between BSA and the chronological subject's age) feature in the context of specific disease detection (binary classification). To do it, we compare this feature with the brain structure volume feature (denoted as $V$). We denote $\text{BSAGE}_{nc}$ as the \bsage without age correction (see Section \ref{section:implementation_detail}). Finally, we propose to take advantage of both \bsage and structure volume biomarker to improve the discriminative capacity of our model.

Table \ref{table:binary_bacc} shows the results of the comparison between different features for different classification problems. The balanced accuracy (BACC) is presented. Other metrics are provided in the appendix. First, we can remark that \bsage (exp. 2, 6) is better than the non corrected version $\text{BSAGE}_{nc}$ (exp. 1, 5) in most classification problems with a large margin (\ie AD, FTD and MS detection). Only in case of PD detection, the version without age correction is better than the corrected version in both in-domain and out-of-domain dataset. Second, we observe that \bsage (exp. 2, 6) is better than structure volume (exp. 3, 7) in MS detection while the structure volume is better in AD detection and SZ detection. In other cases (\ie FTD and PD detection), one feature is better than the other one on in-domain data and worse on out-of-domain data. From this observation, both the apparent brain structure ages and the structure volumes demonstrate discriminative power for different disease detection tasks. Thus, it should be beneficial to combine them for a better discriminative capacity. As a result, the combination of \bsage and structure volume (exp. 4, 8) shows most of the time the best or the second best performance.

\begin{table}[t]
\caption{Ablation study for binary classification tasks. {\color{red} Red}: best result, {\color{blue} Blue}: second best result. The balanced accuracy (BACC) is used to assess the model performance. The results are the average accuracy of 10 repetitions and presented in percentage. We denote $\text{BSAGE}_{nc}$, \bsage and V for respectively \bsage with no age correction, \bsage with age correction and structure volume.}\label{table:binary_bacc}
\footnotesize
\begin{center}
    \begin{minipage}{\textwidth}
        \begin{tabular*}{\textwidth}{@{\extracolsep{\fill}}cccccccc}
        \toprule
        & No. & Features & AD \vs CN & FTD \vs CN & MS \vs CN & PD \vs CN & SZ \vs CN \\
        \midrule
        \multirow{5}{*}{\rotatebox[origin=l]{90}{In-domain}}& & & $N=781$ & $N=547$ & $N=903$ & $N=763$ & $N=614$\\
        
        & 1 & $\text{BSAGE}_{nc}$ & 76.3 & 71.4 & 70.2 & {\color{blue}71.8} & 63.9\\
        & 2 & \bsage & 88.2 & 86.3 & {\color{blue}83.7} & {\color{red}73.5} & 77.3\\
        & 3 & V & {\color{blue}89.1} & {\color{blue}89.4} & 79.4 & 64.8 & {\color{blue}78.2}\\
        & 4 & \bsage + V & {\color{red}91.8} & {\color{red}91.3} & {\color{red}84.6} & 65.7 & {\color{red}81.0}\\
        \midrule
        \multirow{7}{*}{\rotatebox[origin=l]{90}{Out-of-domain}}\\
        & & & $N=2103$ & $N=1273$ & $N=3411$ & $N=1360$ & $N=1310$\\
        & 5 & $\text{BSAGE}_{nc}$ & 62.3 & 63.6 & 79.3 & {\color{red}63.3} & 73.1\\
        & 6 & \bsage & 78.5 & {\color{blue}90.6} & {\color{red}84.3} & 52.8 & 69.0\\
        & 7 & V & {\color{red}86.3} & 90.1 & 71.1 & 58.3 & {\color{blue}76.6}\\
        & 8 & \bsage + V & {\color{blue}86.0} & {\color{red}91.0} & {\color{blue}83.0} & {\color{blue}59.8} & {\color{red}83.2}\\
        \\
        \bottomrule
        \end{tabular*}
    \end{minipage}
\end{center}
\end{table}

\subsubsection{Multi-disease classification}
\label{section:multi_disease_classification}

Table \ref{table:multi_disease_classification} shows the results for the multi-disease classification problem. We estimated the balanced accuracy (BACC), accuracy (ACC) and area under curve (AUC) of our model. We performed classification using the true age (exp. 1, 6) and the predicted subject's age (exp. 2, 7) to confirm that estimating brain age at structure level provides better results than using a global age estimation with real or estimated values. Moreover, these baseline methods enable to estimate the biases present between populations in terms of age. Indeed, there was some bias in age distribution between diseases since for instance the SZ patients were young while compared to the AD patients. Thanks to this analysis, we can observe that \bsage (exp. 3, 8) and V feature (exp. 4, 9) presents a far higher performance than the true age and the predicted subject's age. This suggests that the structure-related information is valuable in classification context. Besides, although the \bsage (exp. 3, 8) presents lower performance than the V feature (exp. 4, 9), the two biomarkers can be mutually used to achieve better classification performance. Indeed, their combination (exp. 5, 10) shows the best performance for all proposed metrics.

\begin{table}[htpb]
\caption{Multi-disease classification results. {\color{red} Red}: best result, {\color{blue} Blue}: second best result. The results are the average accuracy of 10 repetitions and presented in percentage. We denote \bsage and V for \bsage with correction and structure volume.}\label{table:multi_disease_classification}
\begin{minipage}{\textwidth}
    \begin{center}
        \begin{tabular*}{0.6\textwidth}{@{\extracolsep{\fill}}cccccccc}
        \toprule
        & No. & Features & BACC & ACC & AUC \\
        \midrule
        \multirow{5}{*}{\rotatebox[origin=l]{90}{In-domain}} & 1 & True age & 36.8 &  46.4 & 76.8\\
        & 2 & Predicted age & 32.8  & 39.9 & 74.9\\
        & 3 & \bsage & 58.5  & 61.7 & 88.0\\
        & 4 & V & {\color{blue}64.5} & {\color{blue}65.1} & {\color{blue}90.2}\\
        & 5 & \bsage + V & {\color{red}68.7} & {\color{red}69.6} & {\color{red}93.2}\\
        
        \midrule
        \multirow{6}{*}{\rotatebox[origin=l]{90}{Out-of-domain}}\\
        & 6 & True age & 34.0 &  51.4 & 74.3\\
        & 7 & Predicted age & 31.8  & 42.8 & 72.8\\
        & 8 & \bsage & 44.7 & 57.0 & 82.6\\
        & 9 & V & {\color{blue}58.7} & {\color{blue}59.4} & {\color{blue}86.8}\\
        & 10 & \bsage + V & {\color{red}63.3} & {\color{red}66.1} & {\color{red}90.6}\\
        
        \bottomrule
        \end{tabular*}
    \end{center}
\end{minipage}
\end{table}
\subsection{Predicted brain age of different populations}
\label{section:predicted_brain_age_of_different_populations}
\begin{figure}
    \centering
    \includegraphics[width=0.6\textwidth]{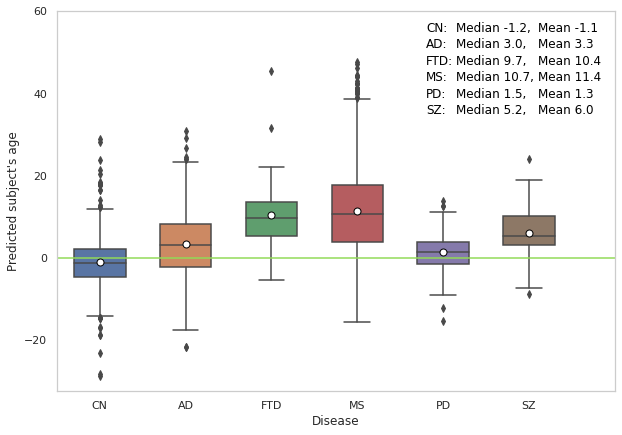}
    \caption{Predicted brain age of different populations in out-of-domain data. The white point presents the position of the mean value.}
    \label{fig:predicted_age_per_population}
\end{figure}

In this section, we compare the predicted brain age between different populations (\ie CN, AD, FTD, MS, PD and SZ). Figure \ref{fig:predicted_age_per_population} summarizes the distribution of predicted brain age for six considered populations. The median and mean predicted brain ages of CN, AD, FTD, MS, PD and SZ are respectively (-1.2, -1.1), (3.0, 3.3), (9.7, 10.4), (10.7, 11.4), (1.5, 1.3) and (5.2, 6.0). First, we observe that the CN class has the mean and median closest to 0 as expected. Second, the BrainAGE of all patient groups is significantly higher than the cognitively normal group ($p<0.0001$ with T-test). Third, PD pathology seems to be closest to healthy people. Indeed, although T1 weighted MRI presents high contrast of grey/white matter, poor contrast may be found in structures related to PD (\eg subthalamic nuclei) \cite{mortezazadeh_imaging_2021}. This may explain the proximity of this class with CN class and the poor performance in PD detection (see Table~\ref{table:binary_bacc}). Fourth, the FTD group presents a more advanced aging process than AD group which is in line with the finding of Lee \ea \cite{lee_deep_2022}. Finally, we found the same magnitude of \bsage for MS (10.7 years) as Cole \ea (about 10.8 years)~\cite{cole_longitudinal_2020}, for AD (3.0 years) as Sendi \ea (2.1 years) \cite{sendi_brain_2021} and for SZ (5.2 years) as Koutsouleris \ea (5.5 years) \cite{koutsouleris_accelerated_2014}.
\subsection{Interpretation of brain structure age gap estimation}
\label{section:interpretation}
\begin{figure}
    \centering
    \includegraphics[width=\textwidth]{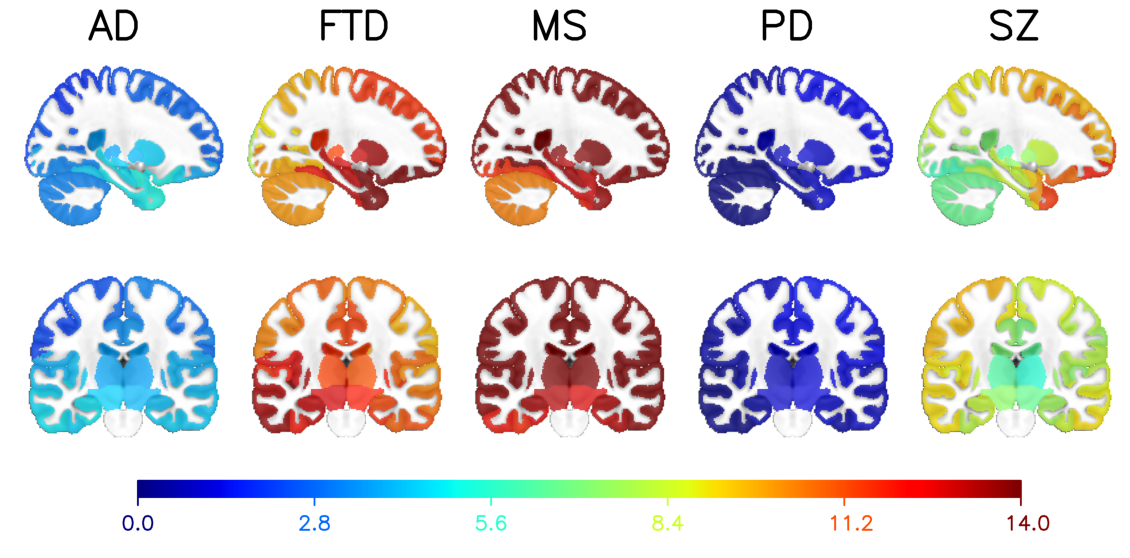}
    \caption{\bsage of different populations in out-of-domain data.}
    \label{fig:avg_map_per_population}
\end{figure}

In this section, we propose to visualize the variation of the age gap between brain structures. The presented results in Figure \ref{fig:avg_map_per_population} correspond to the average \bsage value for each structure on different populations of our out-of-domain datasets. We use the same color bar for all populations to compare the impact of each disease to the aging process.

For the AD group, the region surrounding the hippocampus is highlighted as the most accelerated aging area. This region is well-known to be related to AD \cite{frisoni_clinical_2010, mu_adult_2011, jin_increased_2004, hyman_alzheimers_1984}. For the FTD group, the accelerated aging pattern is mainly located in the temporal and frontal lobes which is in line with current literature~\cite{whitwell_distinct_2009, boeve_advances_2022}. For the MS group, the area with the highest accelerated aging pattern is similar to the finding of Cortese \ea (\ie thalamus and global cortical grey matter)~\cite{cortese_advances_2019}. For the PD group, all regions seem to be close to healthy people as discussed in Section \ref{section:predicted_brain_age_of_different_populations}. Finally, for the SZ group, the prefrontal and medial temporal lobe regions are highlighted which is coherent with several studies~\cite{karlsgodt_structural_2010, delisi_understanding_2006}.
\section{Discussion}
In this work, we proposed an approach to estimate brain age at structure level. We showed that this feature can be used for different purposes. First, it can be directly used to accurately estimate the chronological age. Second, this can be used to compute the \bsage (\ie the difference between brain structure ages and the chronological age). This biomarker presents discriminative patterns which are useful for the multi-disease classification problem (\ie CN \vs AD \vs FTD \vs MS \vs PD \vs SZ).

For the problem of chronological age estimation, we observed that the model accuracy was heavily influenced by different factors: data amount and data augmentation techniques. In our experiments, the model accuracy was consistently improved when the data amount was increased and we did not observe saturation. This suggests that training our framework on more data could yield higher accuracy. In this study, due to the limited training data, we instead applied several data augmentation techniques to improve the generalization capacity of our model. Experimental results showed that the random shift and mixup techniques have a huge impact on our model performance (see Table \ref{table:ablation_choronolical_age_pred}). 
Moreover, the resulting BSA can enable a more accurate subject's age prediction. While our framework was training over a large range of ages (\ie 0 to 95 years old), it achieved higher accuracy when predicting on young population (\ie ABIDE dataset $MAE = 1.91$ years) than on older population (\ie UKBioBank dataset $MAE = 3.87$ years). Our approach outperformed other state-of-the-art methods with 1.17 years MAE lower on the young population and 0.95 year MAE lower on the old population. When evaluating our approach on different disease populations (\ie AD, FTD, MS, PD, SZ), our findings were in line with current knowledge in the literature (see Section \ref{section:predicted_brain_age_of_different_populations} for more details). Finally, we observed that the predicted age distributions are different between diseases, suggesting a discriminative power of our BSA feature.

While most papers used the global BrainAGE to show that a disease can present an accelerated or delayed aging process on a population \cite{franke_brain_2012, koutsouleris_accelerated_2014, franke_estimating_2010}, only a few approaches have proposed to use it for classification \cite{gaser_brainage_2013, franke2014dementia, varzandian_classification-biased_2021, cheng_brain_2021}. Moreover, these studies dedicated to classification had a common limitation. Indeed, it might exist a range of global BrainAGE values that is presented in different populations. In our case, all populations had global BrainAGE values in range [-5, 10] (see Figure \ref{fig:predicted_age_per_population}). When the number of classes is increased, this limitation becomes more challenging. This might explain why existing BrainAGE-based algorithms only address classification problems with a low number of class (\eg binary classification). This raises the need for other features better describing the brain aging process for classification. Thus, we propose to extend the notion of BrainAGE to \bsage. This local feature offers a richer representation of the brain aging process than global BrainAGE estimation. Consequently, this information is important for improving the multi-disease classification as demonstrated in Section \ref{section:multi_disease_classification}.

In addition to improve classification performance, \bsage can be projected into a brain segmentation for visualization purpose. Principal remarks for each population were discussed in Section \ref{section:interpretation}. Overall, it gives some insights about the specific structures impacted by each disease. The main patterns of each disease highlighted by our color maps are coherent with current literature (as discussed in Section \ref{section:interpretation}). This presents an important clinical value of our framework in a real medical context.

Finally, it still exists some limitations in this study. For example, some diseases such as Parkinson's Disease cannot be easily detected using T1 weighted MRI. Future works should focus on the multi-modal input to either accurately estimate brain age or produce a more discriminative \bsage feature. In addition, we use the same CNN architecture to analyze different brain locations. This can be not optimal due to the fact that different brain locations may have a specific set of patterns. An auto-search algorithm to select an optimal architecture for each brain region would be beneficial for further analysis.

\section{Conclusion}
In this paper, we propose to extend the notion of brain age by estimating the brain age at voxel level. This voxelwise brain age map is then used to compute BSA. This biomarker can be used for different purposes. First, it can be used to predict the chronological age of people. The deviation of the predicted age from the subject's age can provide insight about the individual brain status. Second, by subtracting the subject's age from the BSA, we obtain a BSA gap estimation (\ie \bsage). This feature can be mutually used with other biomarkers such as structure volume for disease detection. Finally, this feature can be also visualized to detect brain abnormality in MRI. Such a tool can help clinicians in making more informed decisions.

\section*{Acknowledgement}
This work benefited from the support of the project DeepvolBrain of the French National Research Agency (ANR-18-CE45-0013). This study was achieved within the context of the Laboratory of Excellence TRAIL ANR-10-LABX-57 for the BigDataBrain project. Moreover, we thank the Investments for the future Program IdEx Bordeaux (ANR-10-IDEX-03-02 and RRI "IMPACT"), the French Ministry of Education and Research, and the CNRS for DeepMultiBrain project.

The ADNI data used in the preparation of this manuscript were obtained from the Alzheimer’s Disease Neuroimaging Initiative (ADNI) (National Institutes of Health Grant U01 AG024904). The ADNI is funded by the National Institute on Aging and the National Institute of Biomedical Imaging and Bioengineering and through generous contributions from private partners as well as nonprofit partners listed at: \url{https://ida.loni.usc.edu/collaboration/access/appLicense.jsp}. Private sector contributions to the ADNI are facilitated by the Foundation for the National Institutes of Health (\url{www.fnih.org}). The grantee organization is the Northern California Institute for Research and Education, and the study was coordinated by the Alzheimer’s Disease Cooperative Study at the University of California, San Diego. ADNI data are disseminated by the Laboratory for NeuroImaging at the University of California, Los Angeles. This research was also supported by NIH grants P30AG010129, K01 AG030514 and the Dana Foundation.

The NDAR data used in the preparation of this manuscript were obtained from the NIH-supported National Database for Autism Research (NDAR). This is supported by the National Institute of Child Health and Human Development, the National Institute on Drug Abuse, the National Institute of Mental Health, and the National Institute of Neurological Disorders and Stroke. A listing of the participating sites and a complete listing of the study investigators can be found at \url{http://pediatricmri.nih.gov/nihpd/info/participating_centers.html}.

The ICBM data used in the preparation of this manuscript were supported by Human Brain Project grant PO1MHO52176-11 and Canadian Institutes of Health Research grant MOP- 34996.

The IXI data used in the preparation of this manuscript were supported by the U.K. Engineering and Physical Sciences Research Council (EPSRC) GR/S21533/02 - \url{http://www.brain-development.org/}.

The ABIDE data used in the preparation of this manuscript were supported by ABIDE funding resources listed at \url{http://fcon_1000.projects.nitrc.org/indi/abide/}.

The AIBL data used in the preparation of this manuscript were obtained from the AIBL study of ageing funded by the Common-wealth Scientific Industrial Research Organization (CSIRO; a publicly funded government research organization), Science Industry Endowment Fund, National Health and Medical Research Council of Australia (project grant 1011689), Alzheimer’s Association, Alzheimer’s Drug Discovery Foundation, and an anonymous foundation. See \url{www.aibl.csiro.au} for further details.

The ADHAD, DLBS and SALD data used in the preparation of this article were obtained from \url{http://fcon_1000.projects.nitrc.org} (Mennes M et al., NeuroImage, 2013; Wei D et al., bioRxiv 2017).

Data used in the preparation of this article were also obtained from the MIRIAD database (Malone IB et al., NeuroImage, 2012) The MIRIAD investigators did not participate in analysis or writing of this report. The MIRIAD dataset is made available through the support of the UK Alzheimer's Society (Grant RF116). The original data collection was funded through an unrestricted educational grant from GlaxoSmithKline (Grant 6GKC).

Data used in the preparation of this article were obtained from the Parkinson’s Progression Markers Initiative (PPMI) database (\url{www.ppmi-info.org}). PPMI – a public-private partnership – was funded by The Michael J. Fox Foundation for Parkinson’s Research and funding partners that can be found at \url{https://www.ppmi-info.org/about-ppmi/who-we- are/study-sponsor}

Data collection and sharing for this project was provided by the Cambridge Centre for Ageing and Neuroscience (CamCAN, \url{https://camcan-archive.mrc-cbu.cam.ac.uk/dataaccess/}). CamCAN funding was provided by the UK Biotechnology and Biological Sciences Research Council (grant number BB/H008217/1), together with support from the UK Medical Research Council and University of Cambridge, UK.

Data used in the preparation of this work were obtained from the DecNef Project Brain Data Repository (\url{https://bicr-resource.atr.jp/srpbsopen/}) gathered by a consortium as part of the Japanese Strategic Research Program for the Promotion of Brain Science (SRPBS) supported by the Japanese Advanced Research and Development Programs for Medical Innovation (AMED, Tanaka SC et al., Scientific data, 2021).

TLDNI was funded through the National Institute of Aging, and started in 2010. The primary goals of FTLDNI were to identify neuroimaging modalities and methods of analysis for tracking frontotemporal lobar degeneration (FTLD) and to assess the value of imaging versus other biomarkers in diagnostic roles. The Principal Investigator of NIFD was Dr. Howard Rosen, MD at the University of California, San Francisco. The data are the result of collaborative efforts at three sites in North America. For up-to-date information on participation and protocol, please visit \url{http://memory.ucsf.edu/research/studies/nifd}. Data collection and sharing for this project was funded by the Frontotemporal Lobar Degeneration Neuroimaging Initiative (National Institutes of Health Grant R01 AG032306). The study is coordinated through the University of California, San Francisco, Memory and Aging Center. FTLDNI data are disseminated by the Laboratory for Neuro Imaging at the University of Southern California.

The C-MIND data used in the preparation of this article were obtained from the C-MIND Data Repository (accessed in Feb 2015) created by the C-MIND study of Normal Brain Development. This is a multisite, longitudinal study of typically developing children from ages newborn through young adulthood conducted by Cincinnati Children’s Hospital Medical Center and UCLA and supported by the National Institute of Child Health and Human Development (Contract \#s HHSN275200900018C). A listing of the participating sites and a complete listing of the study investigators can be found at \url{https://research.cchmc.org/c-mind}. The NDAR data used in the preparation of this manuscript were obtained from the NIH-supported National Database for Autism Research (NDAR). NDAR is a collaborative informatics system created by the National Institutes of Health to provide a national resource to support and accelerate research in autism. The NDAR dataset includes data from the NIH Pediatric MRI Data Repository created by the NIH MRI Study of Normal Brain Development. This is a multisite, longitudinal study of typically developing  hildren from ages newborn through young adulthood conducted by the Brain Development Cooperative Group and supported by the National Institute of Child Health and Human Development, the National Institute on Drug Abuse, the National Institute of Mental Health, and the National Institute of Neurological Disorders and Stroke (Contract \#s N01- HD02-3343, N01-MH9-0002, and N01-NS-9-2314, -2315, -2316, -2317, - 2319 and -2320). A listing of the participating sites and a complete listing of the study investigators can be found at \url{http://pediatricmri.nih.gov/nihpd/info/participating_centers.html}.

The NACC database was funded by NIA/NIH Grants listed at \url{https://naccdata.org/publish-project/authors-checklist#acknowledgment}.

This research has been conducted using data from UK Biobank, a major biomedical database. . See \url{https://www.ukbiobank.ac.uk/} for further details.

Data collection has been supported by a grant provided by the French State and handled by the "Agence Nationale de la Recherche," within the framework of the "Investments for the Future" programme, under the reference ANR-10-COHO-002, Observatoire Français de la Sclérose en Plaques (OFSEP)"
\&
"Eugène Devic EDMUS Foundation against multiple sclerosis".

Data was downloaded from the COllaborative Informatics and Neuroimaging Suite Data Exchange tool (COINS; \url{http://coins.mrn.org/dx}) and data collection was funded by NIMH R01MH084898-01A1, “Brain Glutamate and Outcome in Schizophrenia”, PI: J. Bustillo.

Data were provided in part by OASIS :
OASIS-1: Cross-Sectional: Principal Investigators: D. Marcus, R, Buckner, J, Csernansky J. Morris; P50 AG05681, P01 AG03991, P01 AG026276, R01 AG021910, P20 MH071616, U24 RR021382

\bibliographystyle{bibstyles/model1-num-names}
\bibliography{main}

\pagebreak
\clearpage
\section{Annexes}
\begin{longtable}{lcccccc}
\caption{Ablation study for binary classification tasks. {\color{red} Red}: best result, {\color{blue} Blue}: second best result. The accuracy (ACC) is used to assess the model performance. The results are the average accuracy of 10 repetitions and presented in percentage.  We denote $\text{BSAGE}_{nc}$, \bsage and V for \bsage with no age correction, \bsage with age correction and structure volume.}\\
\begin{minipage}{\textwidth}
    \begin{center}
    \begin{tabular*}{\textwidth}{@{\extracolsep{\fill}}cccccccc}
    \toprule
    & No. & Features & AD \vs CN & FTD \vs CN & MS \vs CN & PD \vs CN & SZ \vs CN \\
    \midrule
    \multirow{5}{*}{\rotatebox[origin=l]{90}{In-domain}}& & & $N=781$ & $N=547$ & $N=903$ & $N=763$ & $N=614$\\
    & 1 & $\text{BSAGE}_{nc}$ & 75.8 & 76.4 & 70.9 & {\color{blue}71.3} & 68.9\\
    & 2 & \bsage & 77.1 & 89.8 & {\color{blue}83.5} & {\color{red}73.8} & 81.3\\
    & 3 & V & {\color{blue}89.1} & {\color{blue}93.1} & 79.8 & 65.0 & {\color{blue}81.8}\\
    & 4 & \bsage + V & {\color{red}91.8} & {\color{red}93.8} & {\color{red}84.6} & 66.1 & {\color{red}83.9}\\
    \midrule
    \multirow{7}{*}{\rotatebox[origin=l]{90}{Out-of-domain}}\\
    & & & $N=2103$ & $N=1273$ & $N=3411$ & $N=1360$ & $N=1310$\\
    & 5 & $\text{BSAGE}_{nc}$ & 59.4 & 69.0 & 78.3 & 42.9 & 90.1\\
    & 6 & \bsage & 57.6 & 94.3 & {\color{blue}83.1} & {\color{red}58.2} & {\color{blue}93.1}\\
    & 7 & V & {\color{blue}86.2} & {\color{red}95.4} & 73.4 & 51.6 & 91.3\\
    & 8 & \bsage + V & {\color{red}86.3} & {\color{blue}95.0} & {\color{red}83.5} & {\color{blue}54.9} & {\color{red}94.0}\\
    \\
    \bottomrule
    \end{tabular*}
    \end{center}
\end{minipage}
\end{longtable}

\vspace{10pt}
\noindent
\begin{longtable}{lcccccc}
\caption{Ablation study for binary classification tasks. {\color{red} Red}: best result, {\color{blue} Blue}: second best result. The area under curve (AUC) is used to assess the model performance. The results are the average accuracy of 10 repetitions and presented in percentage.  We denote $\text{BSAGE}_{nc}$, \bsage and V for \bsage with no age correction, \bsage with age correction and structure volume.}\\
    \begin{minipage}{\textwidth}
        \begin{center}
            \begin{tabular*}{\textwidth}{@{\extracolsep{\fill}}cccccccc}
            \toprule
            & No. & Features & AD \vs CN & FTD \vs CN & MS \vs CN & PD \vs CN & SZ \vs CN \\
            \midrule
            \multirow{5}{*}{\rotatebox[origin=l]{90}{In-domain}}& & & $N=781$ & $N=547$ & $N=903$ & $N=763$ & $N=614$\\
            & 1 & $\text{BSAGE}_{nc}$ & 80.8 & 82.3 & 78.0 & {\color{blue}75.7} & 78.8\\
            & 2 & \bsage & 94.8 & 94.6 & {\color{blue}91.5} & {\color{red}79.0} & 88.0\\
            & 3 & V & {\color{blue}95.6} & {\color{blue}95.2} & 87.1 & 71.6 & {\color{blue}88.3}\\
            & 4 & \bsage + V & {\color{red}96.6} & {\color{red}97.2} & {\color{red}93.0} & 72.2 & {\color{red}91.4}\\
            \midrule
            \multirow{7}{*}{\rotatebox[origin=l]{90}{Out-of-domain}}\\
            & & & $N=2103$ & $N=1273$ & $N=3411$ & $N=1360$ & $N=1310$\\
            & 5 & $\text{BSAGE}_{nc}$ & 70.1 & 69.5 & 87.3 & {\color{red}64.9} & {\color{blue}89.8}\\
            & 6 & \bsage & 85.2 & {\color{blue}94.0} & {\color{blue}91.0} & 53.2 & 84.7\\
            & 7 & V & {\color{red}94.0} & 93.9 & 78.9 & 61.8 & 88.9\\
            & 8 & \bsage + V & {\color{blue}93.5} & {\color{red}94.6} & {\color{red}91.2} & {\color{blue}63.3} & \textbf{\color{red}94.2}\\
            \\
            \bottomrule
            \end{tabular*}
        \end{center}
    \end{minipage}
\end{longtable}

\end{document}